\documentclass[11pt,a4paper]{article}

\newcommand{\isready}{} % uncomment for final submission
\newcommand{\ifreadyelse}[2]{\ifdef{\isready}{#1}{#2}}

\newcommand{\withappendix}{} % uncomment for appendix
\newcommand{\ifappendix}[2]{\ifdef{\withappendix}{#1}{#2}}

\usepackage{authblk}
\makeatletter
\newcommand\email[2][]%
   {\newaffiltrue\let\AB@blk@and\AB@pand
      \if\relax#1\relax\def\AB@note{\AB@thenote}\else\def\AB@note{\relax}%
        \setcounter{Maxaffil}{0}\fi
      \begingroup
        \let\protect\@unexpandable@protect
        \def\thanks{\protect\thanks}\def\footnote{\protect\footnote}%
        \@temptokena=\expandafter{\AB@authors}%
        {\def\\{\protect\\\protect\Affilfont}\xdef\AB@temp{#2}}%
         \xdef\AB@authors{\the\@temptokena\AB@las\AB@au@str
         \protect\\[\affilsep]\protect\Affilfont\AB@temp}%
         \gdef\AB@las{}\gdef\AB@au@str{}%
        {\def\\{, \ignorespaces}\xdef\AB@temp{#2}}%
        \@temptokena=\expandafter{\AB@affillist}%
        \xdef\AB@affillist{\the\@temptokena \AB@affilsep
          \AB@affilnote{}\protect\Affilfont\AB@temp}%
      \endgroup
       \let\AB@affilsep\AB@affilsepx
}
\makeatother
\usepackage[hyperref]{emnlp2020}
\usepackage{times}
\usepackage{latexsym}

\usepackage{microtype}

\aclfinalcopy % Uncomment this line for the final submission
\def\aclpaperid{449} %  Enter the acl Paper ID here

\newcommand{\papername}{A Simple and Effective Model for Answering Multi-span Questions}
\newcommand{\drop}{\textsc{DROP}}
\newcommand{\quoref}{\textsc{Quoref}}
\newcommand{\MTMSN}{\textsc{MTMSN}}
\newcommand{\BERTCalc}{\textsc{BERT-Calc}}

\newcommand{\embedding}{$\bm{h}$}
\newcommand{\fone}{F{\scriptsize 1}}
\newcommand{\roberta}{RoBERTa\textsubscript{\tiny{\textsc{LARGE}}}\text{}}
\newcommand{\bert}{BERT\textsubscript{\tiny{\textsc{LARGE}}}\text{}}
\newcommand{\nerd}{NeRd}

\newcommand\jb[1]{\textcolor{blue}{[JB: #1]}}
\newcommand\ag[1]{\textcolor{red}{[AG: #1]}}
\newcommand\es[1]{\textcolor{purple}{[ES: #1]}}
\newcommand\avia[1]{\textcolor{cyan}{[AE: #1]}}
\newcommand\comment[1]{}

\ifreadyelse{
\renewcommand\jb[1]{}
\renewcommand\ag[1]{}
\renewcommand\es[1]{}
\renewcommand\avia[1]{}
}{}

\usepackage{amsfonts} % \mathbb
\usepackage{amsmath} % \begin{align*}
\usepackage{bm} % \bm
\usepackage[capitalise]{cleveref} % \cref
\usepackage{multirow}
\usepackage{graphicx}
\usepackage{makecell}
\graphicspath{ {./images/} }
\usepackage{textcomp} % for PyTorch citation
\usepackage{booktabs} % nice tables

\title{\papername}

\author[1]{\bf Elad Segal}
\author[1]{\bf Avia Efrat}
\author[1]{\bf Mor Shoham}
\author[1,3]{\bf Amir Globerson}
\author[1,2]{\bf Jonathan Berant}

{
\makeatletter
\renewcommand\AB@affilsepx{\quad \protect\Affilfont}
\makeatother
\affil[1]{Tel Aviv University}
\affil[2]{Allen Institute for AI}
\affil[3]{Google Research}}
\email{}

\email{\normalsize \texttt{elad.segal@gmail.com}, \quad \texttt{$\{$aviaefra,morshoham$\}$@mail.tau.ac.il}}
\email{\normalsize \texttt{amir.globerson@gmail.com},\quad \texttt{joberant@cs.tau.ac.il}}

\date{}

\begin{document}

\setlength{\abovedisplayskip}{2pt}
\setlength{\belowdisplayskip}{2pt}

\maketitle

\begin{abstract}
 Models for reading comprehension (RC) commonly restrict their output space to the set of all single contiguous spans from the input, in order to alleviate the learning problem and avoid the need for a model that generates text explicitly.
 However, forcing an answer to be a single span can be restrictive, and some recent datasets also include \emph{multi-span questions}, i.e., questions whose answer is a set of non-contiguous spans in the text.
 Naturally, models that return single spans cannot answer these questions. In this work, we propose a simple architecture for answering multi-span questions by casting the task as a sequence tagging problem, namely, predicting for each input token whether it should be part of the output or not. Our model substantially improves performance on span extraction questions from \drop{} and \quoref{} by 9.9 and 5.5 EM points respectively.
\end{abstract}

\section{Introduction}

The task of reading comprehension (RC), where given a question and context, one
provides an answer, has gained immense attention recently. In most datasets and models \cite{rajpurkar-etal-2016-squad, DBLP:journals/corr/TrischlerWYHSBS16, DBLP:conf/iclr/SeoKFH17,Yu2018QANetCL,47761}, RC is set up as an \emph{extractive}
task, where the answer is constrained to be a single span from the input. This makes learning easier, since the model does not need to generate text
\emph{abstractively}, while still being expressive enough to capture a
large set of questions.

\begin{figure}
    \centering
    \includegraphics[scale=0.425]{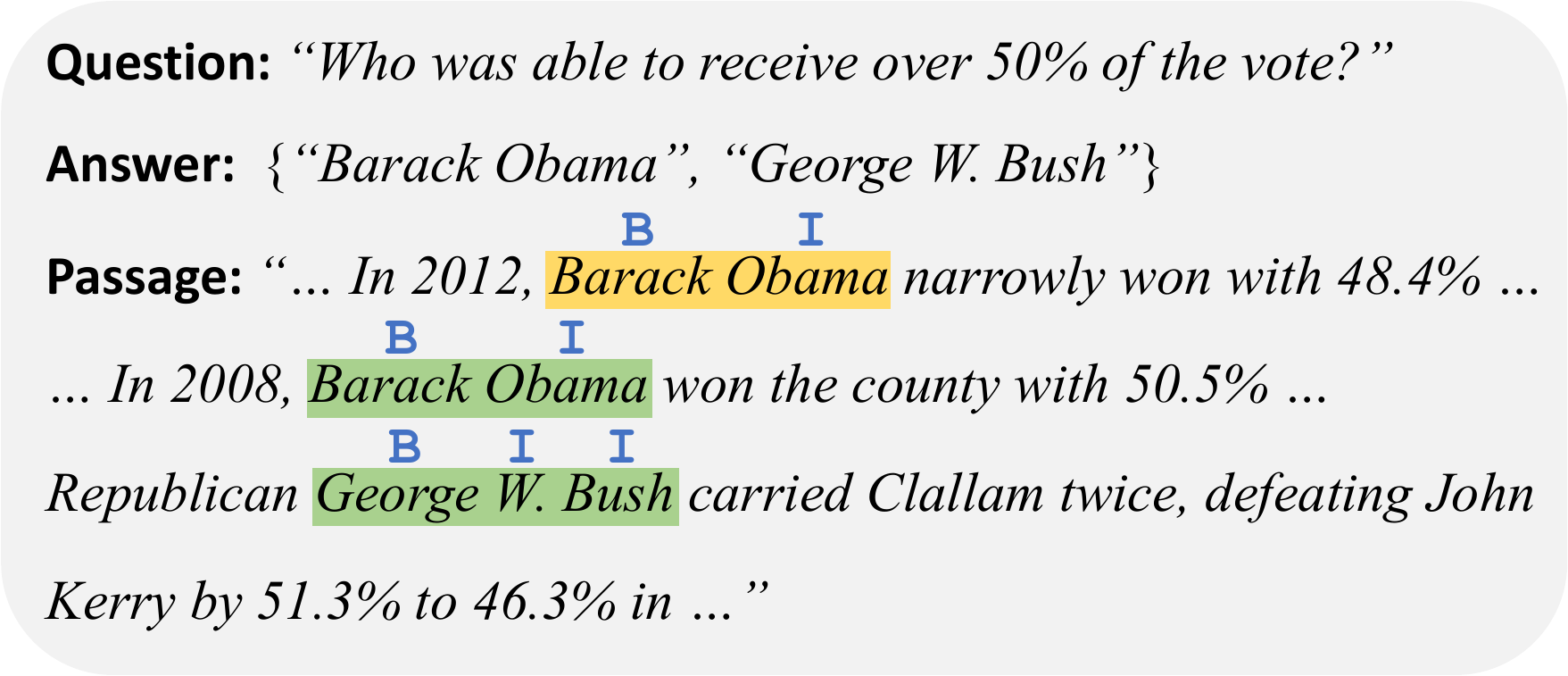}
    \caption{A \emph{multi-span question} from \drop{}, and a \texttt{BIO} tagging for it (\texttt{O} tags omitted). The first occurrence of \emph{Barack Obama} does not answer the question.}
    \label{fig:tag_example}
\end{figure}

However, for some questions, while the answer is indeed extractive, i.e., contained in the input, it is not \emph{a single span}. 
For example, in Figure~\ref{fig:tag_example} the answer includes two people who appear as non-contiguous spans in the context. Existing models \cite{DBLP:conf/iclr/SeoKFH17, Dua2019DROP} are by design unable
to provide the correct answer to such \emph{multi-span questions}.

While most work has largely ignored this issue, recent work has taken
initial steps towards handling multi-span questions. \newcite{hu2019multi}
proposed to predict the number of output spans for each question, and used a non-differentiable inference
procedure to find them in the text, leading to a complex
training procedure. \newcite{Andor_2019} proposed a \texttt{Merge} operation
that merges spans, but is constrained to at most 2 spans. \newcite{Chen_2020} proposed a non-differentiable symbolic approach which outputs programs that compose single-span extractions.

In this work, we propose a simple and fully differentiable architecture for handling
multi-span questions that evades the aforementioned shortcomings, and outperforms prior work.
Similar to \newcite{Yao2013AnswerEA}, who used a linear model over tree-based features, 
we cast question answering as a
\emph{sequence tagging task}, predicting for each token whether it is part of the
answer. At test time, we decode the answer with
standard decoding methods, such as Viterbi.

We show the efficacy of our approach on span-extraction questions from both the \drop{}  \cite{Dua2019DROP} and \quoref{} \cite{Dasigi2019Quoref} datasets. Replacing the single-span architecture with our multi-span approach improves performance by 7.8 and 5.5 EM points respectively. Combining the single-span and multi-span architectures further improves performance by 2.1 EM on \drop{}, surpassing results by other span-extraction methods on both datasets.

\section{Background: Single-span Model}

\paragraph{Setup} Given a training set of question-context-answer triplets
$(q_i, c_i, a_i)_{i=1}^N$, our goal is to learn a function that maps a question-context pair $(q, c)$ to an answer $a$. We briefly review the standard \emph{single-span architecture} for RC \cite{devlin-etal-2019-bert}, which we build upon.

First, we encode the question and context with a pre-trained language
model, such as BERT \cite{devlin-etal-2019-bert}: $\bm{h}=\text{Encoder}([q,c])$, 
where $\bm{h} = (\bm{h}_1, \dots, \bm{h}_m)$ is a sequence of contextualized representations for all input tokens. Then, two parameterized functions (feed-forward networks),
$f_\text{start}\left({\bm{h}}_i\right)$ and
$f_\text{end}\left({\bm{h}}_i\right)$, are used to compute a score for each token,
corresponding to whether that token is the start or the end of the answer. Last,
the start and end probability for each token $i$ is computed as follows:
\begin{align*}
  \bm{p}_i^\text{start} &= \text{softmax}\left(f_\text{start}({\bm{h}_1}), \dots, f_\text{start}({\bm{h}_m)}\right)_i,\\
  \bm{p}_i^\text{end} &= \text{softmax}\left(f_\text{end}({\bm{h}_1}), \dots, f_\text{end}({\bm{h}_m)}\right)_i,
\end{align*}
where both $\bm{p}^\text{start},\bm{p}^\text{end}\in \mathbb{R}^{m \times 1}$. Training is done by minimizing cross entropy of the start and end indices of the gold span, and at test time the answer span is extracted by finding the indices $(s, e)$:
\[
    \left(s, e\right)=\underset{s \leq e}{\arg \max}\ \ \bm{p}_s^{\text{start}}
    \bm{p}_e^{\text{end}}.
\]
\section{Multi-span Model}

\subsection{Span Extraction as Sequence Tagging}
Extracting a variable number of spans from an input text is standard in many natural language processing tasks, such as Named Entity Recognition (NER)  and is commonly cast as a sequence tagging problem  \cite{DBLP:journals/corr/cmp-lg-9505040}.
Here we apply this approach to multi-span questions.

Our model uses the same contextualized representations \embedding, but rather than
predicting start and end probabilities, it outputs a probability distribution over a set of tags for each token. We experiment with two tagging schemes. First, the well-known \texttt{BIO} tagging \cite{Sang2000TransformingAC,huang2015bidirectional}, in which \texttt{B} denotes the first token of an output span, \texttt{I} denotes subsequent tokens in a span, and \texttt{O} denotes tokens that are not part of an output span. In addition, we experiment with a simpler \texttt{IO} tagging scheme, where words are tagged as either part of the answer (\texttt{I}) or not (\texttt{O}). Formally, given a tagging scheme with $\lvert S \rvert$ tags ($\lvert S \rvert=3$ for \texttt{BIO} and $\lvert S \rvert=2$ for \texttt{IO}), for each of the $m$ tokens, the probability for the tag of the $i$-th token is
\begin{equation} \label{eq:tag_prob}
      \bm{p}_{i} = \text{softmax}(f(\bm{h}_{i}))
\end{equation}
where $\bm{p} \in \mathbb{R}^{m \times \lvert S \rvert}$, and $f$ is a parameterized function with $\lvert S \rvert$ outputs. 

\subsection{Training}
Assume each answer $a$ is a set of strings, where each string corresponds to a span in the input. 
We would like to train our model to predict the correct output for this set of spans. When the answer spans appear only once in the input, this is simple, since the ground-truth tagging is immediately available. However, there are many cases where a given answer span appears multiple times in the input. We next explain how to address this. 

To illustrate, consider the following simple example (assume a \texttt{BIO} scheme). Given the input ``X Y Z Y Z" and the correct multi-span answer \{``X", ``Z"\}, there are three possible gold taggings: $\mathtt{B\,O\,B\,O\,B}$, $\mathtt{B\,O\,B\,O\,O}$, and $\mathtt{B\,O\,O\,O\,B}$.
Thus, the ground-truth \texttt{BIO} cannot be determined unambiguously in this case. Figure~\ref{fig:tag_example} illustrates this issue with a real example from \drop{}.\footnote{In \quoref{}, the indices of the gold answer spans are explicitly given, so a single gold tagging can be defined.} 

To tackle the above issue, we enumerate over the set of all \emph{possibly-correct taggings}, $\mathcal{T}$, where given a multi-span answer $a$, a possibly-correct tagging is one in which all gold answer spans are tagged as such at least once.\footnote{While $\lvert \mathcal{T} \rvert$ can grow exponentially with the number of spans in an answer, in practice $\lvert \mathcal{T} \rvert$ is at most 1000 for 99.66\% of the examples of \drop, and so we can enumerate over $\mathcal{T}$ directly in these cases. In the other 0.34\%, we take a single tagging that marks all occurrences of the answer spans.}
We train our models by maximizing the marginal probability of all possibly-correct taggings:
\[
\log p(\mathcal{T} \mid \bm{h}) =
\log \sum_{\bm{T} \in \mathcal{T}} \left(
\prod_{i=1}^{m}\bm{p}_{i}[\bm{T}_{i}] \right),
\]
where $\bm{p}_{i}[\bm{T}_{i}]$ (see Eq.~\eqref{eq:tag_prob}) is the probability the model assigns to token $i$ having the tag $\bm{T}_i$. 
The loss is minimized when $\bm{p}$ gives probability $1.0$ to one of the possibly-correct taggings in $\mathcal{T}$.

\subsection{Decoding Spans from a Tagging}

At test time, given predicted tag probabilities $\bm{p}$, we would like to find the most likely tagging $\hat{\bm{T}}$.  Let $\mathcal{V}$ be the set of all valid taggings. We wish to find:
\[
\hat{\bm{T}}=\underset{\bm{T} \in
\mathcal{V}}{\arg \max} \prod_{i=1}^{m}\bm{p}_{i}[\bm{T}_{i}].
\]
For \texttt{BIO} tags, the set $\mathcal{V}$ comprises all taggings that don't include an \texttt{I} after an \texttt{O}, and the maximization problem can be solved in linear time using Viterbi decoding
\cite{DBLP:journals/tit/Viterbi67}
as in \citet{Yao2013AnswerEA, Mehta2018TowardsSL}.
For \texttt{IO} tags, all taggings are valid, and maximization is done by predicting the tag with highest probability in each token independently. 
Because answer spans are (practically) never adjacent in RC, an \texttt{IO}-tagging produces a set of spans by choosing all maximal spans that are contiguously tagged with \texttt{I}.

\section{``Multi-Head" Models}

Some RC datasets contain questions where the output is not necessarily a span.
For example, in \drop{}, the answer to some questions is a number that is not in the text, but can be computed by performing arithmetic operations. To handle such cases, many models \cite{Dua2019DROP, hu2019multi} employ a \textit{multi-head architecture}.
In these models, each \textit{head} $z$ is a small module that  takes the contextualized representations \embedding{} as input and computes a probability distribution over answers $p_z(a \mid q, c) = p_z(a \mid \bm{h})$. 
For example, in \newcite{hu2019multi}, there are two heads that output spans, and three heads that output numbers. To determine which head to use for each question, an additional module is trained: $p_\text{head}(z \mid q, c) = p_\text{head}(z \mid \bm{h})$. Thus, the model probability for an answer is:
\[
p(a \mid q, c) = \sum_z p_\text{head}(z \mid q, c)\cdot p_z(a \mid q, c).
\]

With this architecture, we can seamlessly integrate our multi-span approach into
existing RC models. Specifically, a model can include both a single-span head and a multi-span head, dynamically deciding which span extraction method to utilize based on the input.

\section{Empirical Evaluation}
\label{sec:experiments}

\paragraph{Experimental setup}
As an encoder, we use the Hugging Face implementation of \roberta{} \cite{Wolf2019HuggingFacesTS, liu2019roberta}, which produces the representations $\bm{h}$. For \drop{}, we add the arithmetic and count heads from \newcite{Dua2019DROP} to handle non-span questions. Full details of the experimental setup are in Appendix~\ifappendix{\ref{app:setup}}{A}.

\subsection{Results}

\begin{table*}[ht]{
\centering
\footnotesize
\begin{tabular}{|l|cc|cc|cc|cc|cc|cc|}
\hline
\multicolumn{1}{|c|}{\multirow{3}{*}{Model}} & \multicolumn{6}{c|}{\drop{}} & 
\multicolumn{6}{c|}{\quoref{}} \\
\cline{2-13} &
\multicolumn{2}{c|}{All Spans} &
\multicolumn{2}{c|}{Multi-Span} & 
\multicolumn{2}{c|}{Single-Span} &
\multicolumn{2}{c|}{All Spans} &
\multicolumn{2}{c|}{Multi-Span} & 
\multicolumn{2}{c|}{Single-Span} \\
\multicolumn{1}{|c|}{}& EM & \fone{} & EM & \fone{} & EM & \fone{} & EM & \fone{} & EM & \fone{} & EM & \fone{} \\ \hline
\BERTCalc{}  & 69.1 & 78.9 & 6.2 & 47.0 & 79.8 & 84.3 & - & - & - & - & - & -\\
\MTMSN{}  & 69.7 & 79.9 & 25.1 & 62.8 & 77.5 & 82.8 & - & - & - & - & - & - \\ 

\nerd{}  & 73.2 & 81.3 & 51.3 & 77.6 & 76.2 & 81.8 & - & - & - & - & - & - \\ 

Coref\roberta & - & - & - & - & - & - & 74.9 & 81.7 & 38.8 & 65.9 & 78.7 & 83.3 \\ \hline

TASE{\scriptsize BIO} + SSE (\bert)  & 76.4 & 83.9 & 53.6 & 76.9 & 80.2 & 85.1 & 75.8 & 81.1 & 52.5 & 76.7 & 78.2 & 81.6 \\

TASE{\scriptsize BIO} + SSE  & 79.7 & 87.1 & 56.3 & 79.9 & 83.6 & 88.3 & 79.0 & 84.2 & \textbf{59.7} & \textbf{80.0} & 80.9 & 84.6 \\

TASE{\scriptsize IO} + SSE  & \textbf{80.5} & \textbf{87.8} & \textbf{58.5} & \textbf{80.7} & \textbf{84.2} & \textbf{89.0} & \textbf{79.4} & 84.8 & 57.9 & 79.2 & \textbf{81.6} & \textbf{85.4} \\ 
\hline

TASE{\scriptsize BIO}  & 77.9 & 85.5 & 56.6 & 79.3 & 81.5 & 86.6 & 78.9 & 84.6 & 56.6 & 77.5 & 81.2 & 85.3 \\

TASE{\scriptsize IO}  & 78.4 & 86.8 & 56.8 & 79.8 & 82.1 & 88.0 & \textbf{79.4} & \textbf{84.9} & 59.3 & \textbf{80.0} & 81.4 & \textbf{85.4} \\ 

\hline
SSE  & 70.6 & 80.2 & 0.0 & 37.6 & 81.5 & 86.7 & 73.9 & 80.7 & 0.0 & 37.4 & 81.4 & 85.0 \\ \hline
TASE{\scriptsize BIO}, \textsc{NoMarg.} & 76.2 & 85.0 & 54.7 & 79.0 & 79.8 & 86.1 & - & - & - & - & - & - \\ \hline

\end{tabular}
\caption{Development set results on \drop{} and \quoref{} questions whose answer is a span (or list of spans).}
\label{tab:main-results}
}
\end{table*}

Table~\ref{tab:main-results} shows development set results on the \emph{span-extraction questions} of \drop{} \cite{Dua2019DROP} and \quoref{} \cite{Dasigi2019Quoref}. We compare the previous best-performing multi-span models to a combination of our multi-span architecture (TASE: TAg-based Span Extraction) with the traditional single-span extraction (SSE), as well as to each separately.

\paragraph{Comparison to previous models}

For a fair comparison with prior work on \drop{}, we also train our model initialized with \bert{}, as all prior work used it as an encoder.
On \drop{}, TASE{\scriptsize BIO}+SSE (\bert) outperforms all prior models that handle multi-span questions, improving by at least 3.2 EM points. On multi-span questions, we dramatically improve performance over \BERTCalc{} and \MTMSN{}, while obtaining similar performance to \nerd{}.
On \quoref{}, compared to Coref\roberta{} \cite{Ye2020CoreferentialRL} which uses the same method as \MTMSN{} for multi-span extraction, we achieve a substantial improvement of over 20 EM on multi-span questions and an improvement of 4.5 EM and 3.2 \fone{} on the full development set, where the best results are achieved when using solely our multi-span architecture with \texttt{IO}-tagging.

\paragraph{Comparing span extraction architectures}
Table~\ref{tab:main-results} also shows that in both \drop{} and \quoref{}, \emph{replacing} the single-span extraction architecture with our multi-span extraction results in dramatic improvement in multi-span question performance, while single-span question performance is either maintained or improved. Furthermore, although combining both architectures tends to yield the best overall performance,\footnote{As single-span questions outnumber multi-span questions in \drop{} and \quoref{} 1:7 and 1:10 respectively, the overall span performance (``All Spans") gives a much larger weight to single-span performance.} the improvement over using only our multi-span architecture is not substantial, suggesting that the multi-span architecture may be used by itself as a general span extraction method.

\paragraph{Effects of tagging scheme}
Overall, the results are quite similar for the \texttt{BIO} and \texttt{IO} schemes. The slight advantage of \texttt{IO} could perhaps be explained by the fact that the model no longer requires distinguishing between \texttt{B} and \texttt{I}, in the presence of powerful contextualized representations.

\paragraph{Effect of marginalization}
To check whether marginalizing over all possibly-correct taggings is beneficial, we ran TASE{\scriptsize BIO} in a setup where only a single tagging is considered, namely where all occurrences of a gold answer span are tagged. Table~\ref{tab:main-results} shows that this indeed leads to a moderate drop of up to 2 points in performance.

\paragraph{Test set results}
We ran TASE{\scriptsize IO} on the \quoref{} test set. Our model obtains 79.7 EM and 86.1 \fone{}, an improvement of 3.9 EM points and 3.3 \fone{} points over the state-of-the-art Coref\roberta.
On \drop{}, our TASE{\scriptsize IO}+SSE model achieves 80.4 EM and 83.6 \fone{} on the entire test set (including non-span questions).

We note that the top 10 models on the \drop{} leaderboard (as of September 15, 2020) have all incorporated our multi-span head using our code base which has been public for a while.

\subsection{Analysis}

\begin{figure}
    \centering
    \includegraphics[scale=0.4]{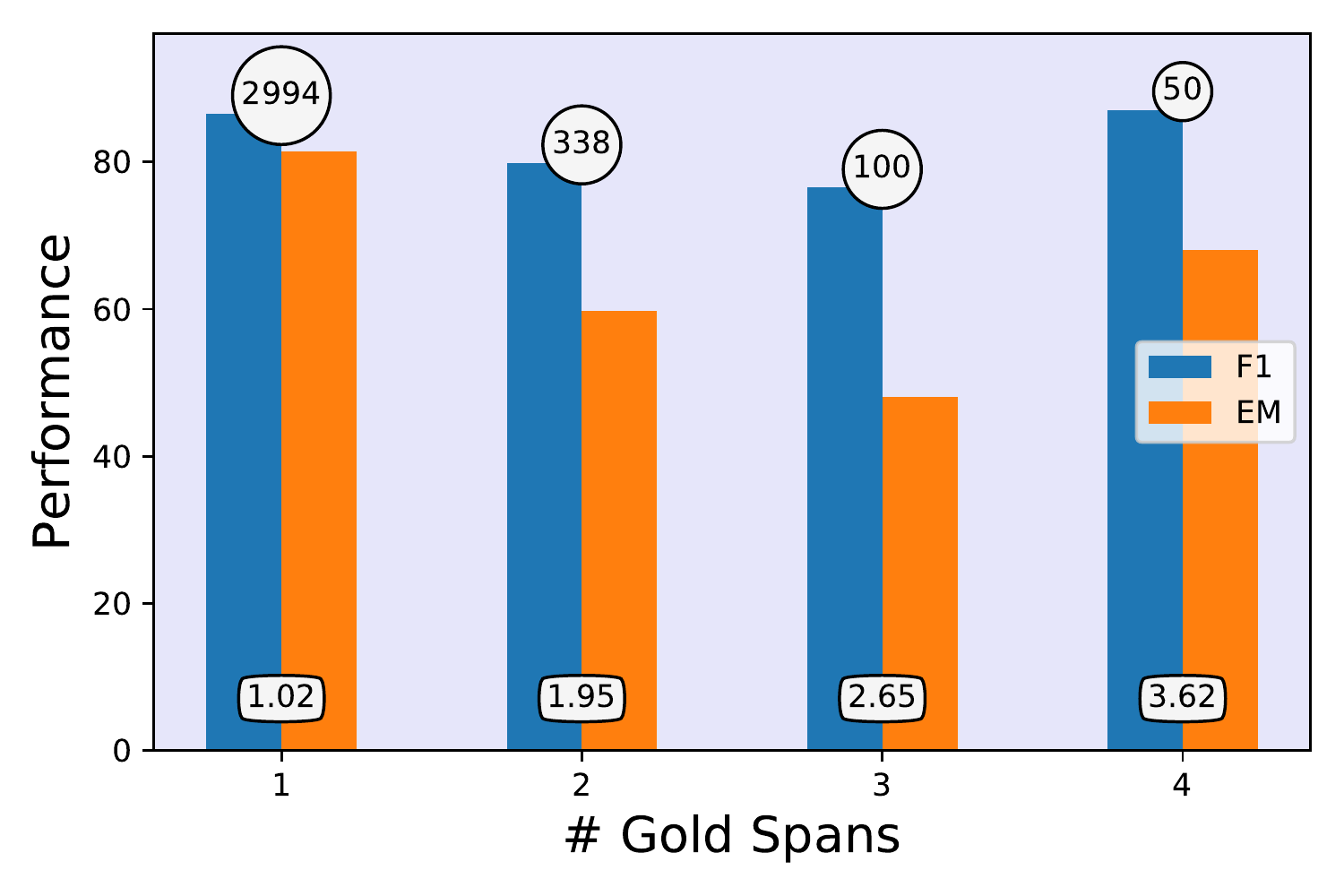}
    \caption{\drop{} Performance of TASE{\scriptsize BIO} by number of spans in the gold answer. Labels at the bottom indicate the average number of predicted spans. Circles at the top are the number of examples. These same trends are observed in \quoref{} as well.}
    \label{fig:n_gold_spans}
\end{figure}

Figure~\ref{fig:n_gold_spans} shows that in both \drop{} and \quoref{} the performance of TASE{\scriptsize BIO} decreases only moderately as the number of gold spans increases. This shows relative robustness to the number of answer spans. In addition, we can see that our architecture is quite accurate in predicting the correct number of spans, with a tendency for under-estimation.

We analyzed the performance of the $p_\text{head}$ module in TASE{\scriptsize BIO}+SSE. A non-multi-span head is selected erroneously for 3.7\% and 7.2\% of the multi-span questions in \drop{} and \quoref{} respectively. The multi-span head is selected for 1.2\% and 1.5\% of the single-span questions in \drop{} and \quoref{} respectively. However, this is reasonable as the multi-span head is capable of answering single-span questions as well, and indeed it returned a single span in 45\% of these cases on both datasets.

We manually analyzed errors of TASE{\scriptsize BIO}+SSE on \drop{}, and  detected 3 main failure cases: (1)
questions where the answer is a span, but requires some numerical computation internally, (2) questions where the number of output spans is explicitly mentioned in the question but is not followed by the model, and (3) questions where a single contiguous span is unnecessarily split into two shorter spans. An example for each case is given in Appendix~\ifappendix{\ref{app:error_analysis}}{B}.
\section{Conclusion}
In this work, we cast the task of answering multi-span questions as a sequence tagging problem, and present a simple corresponding multi-span architecture. 
We show that replacing the standard single-span architecture with our multi-span architecture dramatically improves results on multi-span questions, without harming performance on single-span questions, leading to state-of-the-art results on \quoref{}. In addition, integrating our multi-span architecture into existing models further improves performance on \drop{}, as is evident from the leading models on \drop{}'s leaderboard.
Our code can be downloaded from \href{\ifreadyelse{https://github.com/eladsegal/tag-based-multi-span-extraction}{https://anonymized}}{\ifreadyelse{https://github.com/eladsegal/tag-based-multi-span-extraction}{https://anonymized}}.
\ifreadyelse{\section*{Acknowledgements}
This research was partially supported by The Israel Science Foundation grants 942/16 and 1186/18, The Yandex Initiative for Machine Learning and the European Research Council (ERC) under the European Union Horizons 2020 research and innovation programme (grant ERC DELPHI 802800).}{}

\bibliographystyle{acl_natbib}
\bibliography{emnlp2020}

\ifappendix{
\appendix
\section*{Appendix for ``\papername''}
\section{Experimental Setup}
\label{app:setup}
We experiment with model variations that use either SSE, TASE, or their combination as a multi-head model. For \drop{}, we additionally use \textit{arithmetic} and \textit{count} heads based on \cite{Dua2019DROP, nabert+}.
Our model is implemented with PyTorch \cite{NEURIPS2019_9015} and AllenNLP \cite{Gardner2017AllenNLP}.
For $f$ in \ifappendix{Eq.~\eqref{eq:tag_prob}}{Eq. (1)} we use a 2-layer feed-forward network with ReLU activations and $\lvert S \rvert$ outputs.
We use the Hugging Face implementation of \roberta{} \cite{Wolf2019HuggingFacesTS, liu2019roberta} as the encoder in our model. 5\% of \drop{} and 30\% of \quoref{} are inputs with over 512 tokens.
Due to \roberta{}'s limitation of 512 positional embeddings, we truncate inputs by removing over-flowing tokens from the passage, both at train and test time. We discard 3.87\% of the training examples of \drop{} and 5.05\% of the training example of \quoref{}, which are cases when the answer cannot be outputted by the model (due to a dataset error, or truncation of the correct answer span).
For training, the BertAdam\footnote{\url{https://github.com/huggingface/transformers/blob/694e2117f33d752ae89542e70b84533c52cb9142/README.md\#optimizers}} optimizer is used with default parameters and learning rates of either $5\times10^{-6}$ or $10^{-5}$. Hyperparameter search was not performed. We train on a single NVIDIA Titan XP with a batch size of 2 and gradient accumulation of 6, resulting in an effective batch size of 12, for 20 epochs with an early-stopping patience of 10. The average runtime per epoch is 3.5 hours.
Evaluation was performed with the official evaluation scripts of \drop{} and \quoref{}.
Our full implementation can be found at \href{\ifreadyelse{https://github.com/eladsegal/tag-based-multi-span-extraction}{https://anonymized}}{\ifreadyelse{https://github.com/eladsegal/tag-based-multi-span-extraction}{https://anonymized}}.
\section{Failure Cases Examples}
\label{app:error_analysis}
Table~\ref{tab:failurecases} contains example failure cases of TASE{\scriptsize BIO}+SSE on \drop{}.

\begin{table*}[ht]
\centering
\footnotesize
\begin{tabular}{@{}p{4.5cm} p{5cm} p{2.2cm} p{2.8cm} @{} }
\toprule
\emph{\textbf{Question}} & \emph{\textbf{Excerpt from Context}} & \emph{\textbf{Gold Answer}} & \emph{\textbf{Prediction}}  \\ \midrule 
\addlinespace

Which two nationalities have the same number of immigrants in Bahrain? &
Indians, 125,000 Bangladeshis, 45,000 Pakistanis, 45,000 Filipinos, and 8,000 Indonesians &
\{``Filipinos", \newline ``Pakistanis"\} &
\{``Filipinos", \newline ``Pakistanis", \newline ``Indonesians"\}
\\ \addlinespace \midrule \addlinespace

What event happened first, Spain losing all territories it had gained since 1909, or the Spanish retaking their major fort at Monte Arruit? &
August 1921, Spain lost all the territories it had gained since 1909 [...] By January 1922 the Spanish had retaken their major fort at Monte Arruit &
\{``Spain lost all the territories it had gained since 1909"\} &
\{``August 1921, Spain lost all the", ``territories it had gained since 1909"\}
\\\addlinespace \bottomrule
\end{tabular}
\caption{Example failure cases of TASE{\scriptsize BIO}+SSE on \drop{}. The first answer exhibits a lack of numeric reasoning and ignores the expected number of spans stated in the question. The second splits a correct span into two spans.}
\label{tab:failurecases}
\end{table*}

}{}

\end{document}